\makeatletter\def\graphicscache@inhibit{true}\makeatother

\documentclass[letterpaper, 10 pt, conference]{ieeeconf}  %

\IEEEoverridecommandlockouts                              %

\usepackage{amsmath} %
\usepackage{amssymb}  %
\usepackage{nccmath}
\usepackage{graphicx}
\usepackage{booktabs}

\usepackage{tikz}
\usepackage{pgfplots}
\usetikzlibrary{fit,shapes.callouts,shapes.geometric,backgrounds,positioning,arrows.meta}
\graphicspath{{./figures/}}

\usepackage{tensor}

\usepackage[hidelinks]{hyperref}
\usepackage[capitalize]{cleveref}
\usepackage[style=ieee,hyperref,natbib=true,backend=bibtex,firstinits,doi=false,%
mincitenames=1,maxcitenames=2,maxbibnames=99,sorting=none,terseinits=false,hyperref=true]{biblatex}
\bibliography{references_2022.bib}
\renewbibmacro*{bbx:savehash}{}%
\AtEveryBibitem{\clearfield{pages}\clearlist{publisher}\clearlist{organization}} %
\defbibheading{bibliography}[\bibname]{\section*{References}}

\usepackage{lipsum}
\usepackage[export]{adjustbox}
\newcommand{\T}[2]{\ensuremath{\tensor[^{#2}]{T}{_{#1}}}}
\newcommand{\Ti}[3]{\ensuremath{\tensor*[^{#2}]{T}{^{#3}_{#1}}}}

\newcommand{\Ii}[3]{\ensuremath{\tensor*[^{#2}]{I}{^{#3}_{#1}}}}
\newcommand{\IT}[2]{\ensuremath{\tensor[^{#2}]{I}{_{#1}}}}
\newcommand{\f}[2]{\ensuremath{\tensor[^{#2}]{f}{_{#1}}}}

\usepackage{tikz}
\usetikzlibrary{arrows}
\usetikzlibrary{positioning,calc}
\usetikzlibrary{decorations.pathreplacing}
\usetikzlibrary{decorations.markings}
\usetikzlibrary{decorations.pathmorphing}
\usetikzlibrary{calligraphy}
\usetikzlibrary{fit}
\usetikzlibrary{shapes.arrows}
\usetikzlibrary{shapes.geometric}
\usetikzlibrary{matrix}

\pgfdeclarelayer{background}
\pgfdeclarelayer{foreground}
\pgfsetlayers{background,main,foreground}

\usepackage{pgfplots}

\makeatletter

\tikzdeclarecoordinatesystem{rel}{%
	\tikzset{cs/.cd,x=0pt,y=0pt,#1}%
	\pgfpointlineattime{(\tikz@cs@x / 100)}%
	{\pgfpointanchor{\tikz@pp@name{\tikz@cs@node}}{south west}}%
	{\pgfpointanchor{\tikz@pp@name{\tikz@cs@node}}{south east}}%
	\edef\tikz@cs@x{\the\pgf@x}%
	\pgfpointlineattime{(\tikz@cs@y / 100)}%
	{\pgfpointanchor{\tikz@pp@name{\tikz@cs@node}}{south west}}%
	{\pgfpointanchor{\tikz@pp@name{\tikz@cs@node}}{north west}}%
	\pgfpoint{\tikz@cs@x}{\pgf@y}%
}

\makeatother

\IfFileExists{graphicscache.sty}{\usepackage[dpi=1800]{graphicscache}}

\title{\LARGE \bf
VR Facial Animation for Immersive Telepresence Avatars
}

\author{Andre Rochow, Max Schwarz, Michael Schreiber, and Sven Behnke%
\thanks{All authors are with the Autonomous Intelligent Systems group of University of Bonn, Germany; {\tt rochow@ais.uni-bonn.de}}%
\thanks{$^{1}$\url{https://www.xprize.org/prizes/avatar}}
}

\begin{document}

\maketitle
\thispagestyle{empty}
\pagestyle{empty}

\begin{abstract}
VR Facial Animation is necessary in applications requiring clear view of the face, even though a VR headset is worn.
In our case, we aim to animate the face of an operator who is controlling our robotic avatar system.
We propose a real-time capable pipeline with very fast adaptation for specific operators.
In a quick enrollment step, we capture a sequence of source images from the operator without the VR headset which contain all the important operator-specific appearance information. During inference, we then use the operator keypoint information extracted from a mouth camera and two eye cameras to estimate the target expression and head pose, to which we map the appearance of a source still image.
In order to enhance the mouth expression accuracy, we dynamically select an auxiliary expression frame from the captured sequence.
This selection is done by learning to transform the current mouth keypoints into the source camera space, where the alignment can be determined accurately.
We, furthermore, demonstrate an eye tracking pipeline that can be trained in less than a minute, a time efficient way to train the whole pipeline given a dataset that includes only complete faces, show exemplary results generated by our method, and discuss performance at the ANA Avatar XPRIZE semifinals.
\end{abstract}

\section{Introduction}
\begin{tikzpicture}[remember picture,overlay]
\node[anchor=north,align=center,font=\sffamily\small,yshift=-0.4cm] at (current page.north) {%
	Published in IEEE/RSJ International Conference on Intelligent Robots and Systems (IROS) 2022.};
\end{tikzpicture}%
Virtual Reality (VR) Facial Animation is a challenging sub-problem of facial animation, which aims to generate realistic face images.
It has many applications, such as in computer games, training data generation, virtual reality, and video compression.
VR allows virtual immersion in another world. 
When interacting virtually with a VR user, it is often desirable to perceive all their characteristics and facial expressions. However, VR headsets occlude a large area of the face, which makes direct face capture impossible.

Especially in recent years, deep learning techniques have made great progress in generative modeling~\citep{gan} and rendering~\citep{nerf}. In this work, we demonstrate a deep learning-based approach to the VR Facial Animation problem.
Our main motivation is the ANA Avatar XPRIZE Competition$^1$, where judges interact through avatar systems developed by the participant teams. Here, a specific imposition for a facial animation system is that adaption to the operator has to be finished in one hour. This time has to be shared with the operator training time, so that available time for adaptations is even shorter.
Our avatar robotic system allows an operator to directly perceive the world through the eyes of the avatar. We animate the operator on a display which is attached via a 6\,DoF arm to the avatar. For more information about our avatar system we refer to \citet{schwarz2021nimbro}.
\begin{figure}[t]
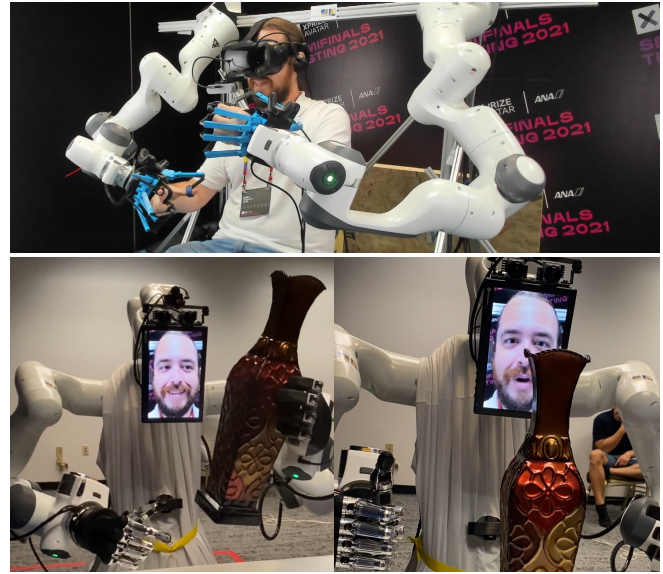

	\centering
	\includegraphics[width=\linewidth,clip,trim=200 150 0 80]{images/operator.jpg} \\
	\vspace{0.2em}
	\includegraphics[width=0.5\linewidth,clip,trim=0 10cm 1cm 6cm]{images/cori_run_avatar1.png}\hfill
	\includegraphics[width=0.5\linewidth,clip,trim=0 13cm 1cm 3cm]{images/cori_run_avatar2.png}
	\caption{Operator interacting through the NimbRo Avatar system with a human recipient at the ANA Avatar XPRIZE semifinals. Top: Operator at remote site.
		Bottom: Two different facial expressions of the operator animated with our method. See also the supplementary video for an animated version.}
	\label{fig:teaser}\vspace{-3ex}
\end{figure}
Inspired by the method of \citet{fom}, our pipeline is trained with the VoxCeleb~\cite{vox} dataset to animate a face
from a source frame, driven by keypoints from a target frame.
We demonstrate how the pipeline can be adapted to generalize to the VR facial animation sub-problem. To enhance the modeling capabilities, we select a third frame, i.e. the expression frame.
Using this frame, we directly embed operator-specific information, which is captured by a camera attached below the HMD (mouth camera), into the animation process. 
We propose an inference pipeline that dynamically and automatically chooses the optimal expression frame via keypoint-driven image retrieval.
Our pipeline runs in real time and can be adapted to an unknown operator with only 15\,min preprocessing.

In addition to the full real-time VR facial animation pipeline, our contributions include:
\begin{enumerate}
    \item A method allowing fast adaptation to new operators,
	\item a training regime that allows offline training on the well-known VoxCeleb~\citep{vox} dataset,
	\item a fast approximate approach to solve the alignment problem between facial image sequences captured with and without a HMD, allowing retrieval of matching frames,
	\item an efficient eye tracking method for challenging camera perspectives, trainable in under a minute, and
	\item methods encouraging temporal continuity at inference.
\end{enumerate}

\section{Related Work}
\paragraph{Facial Image Animation}
Facial Image Animation is a long-standing problem in computer graphics and aims to generate new facial images with controllable expressions.
\citet{x2face} propose a method that learns the sampling coordinates from a source image to an embedded image and then from the embedded image to a driving frame using a separate network. Whereas other warping-based methods exist~\cite{monkey_net,fom,zhao2021sparse,fewshotvid2vid}, there are also indirect approaches~\cite{stargan, ganimation,zakharov2019few} that rely on generative modeling to perform facial expression synthesis. \citet{zakharov2019few} learn an embedding vector from few source images which then modulate a keypoint-driven generator network via Adaptive Instance Normalization (AdaIN)~\cite{ada_in}, which has demonstrated to be especially well-suited to perform style transfer.
For VR facial animation, those approaches, however, must be adapted to work with partially occluded driving frames, e.g. captured from a mouth camera.

\paragraph{Keypoint-Driven Motion Transfer}
\citet{monkey_net} use deep neural networks to decouple appearance and motion information. They combine the appearance extracted from a source image and the motion derived from the driving video. Their pipeline is separated into a keypoint detector, a dense motion network, and a generator network. Based on this architecture,
\citet{fom} encode motions based on keypoints and introduce local affine transformations that allow to model a larger family of transformations.
More recently, \citet{zhao2021sparse} proposed an encoder/decoder dense motion network that employs AdaIn~\cite{ada_in} to transfer source face keypoint geometry to the encoder and driving face keypoint geometry to the decoder. They separate the dense motion network into a global branch and multiple local branches that have a limited visibility---to focus on generating a more accurate motion for the eyes and the mouth area.
Furthermore, they investigate how to improve the temporal alignment of the keypoint detector as proposed by \citet{face_alignment}, which is also the default choice in our approach.
Our method is based on \citet{fom}, but faces a much more difficult problem where the driving frame has occlusion caused by the VR headset and is captured from a perspective different from the source frame.
To address these issues, we forgo using local affine transformations~\cite{fom} and rely on a keypoint-based image retrieval to simulate the lower face region more precisely.

\paragraph{Virtual Reality Facial Animation}
VR Facial Animation is a special case of facial animation where large parts of the face are occluded by an HMD. \citet{facevr} use image retrieval to obtain the most similar source views. They use blending to combine the retrieved images to a photo-realistic output. Several methods~\cite{vr_facial,lombardi2018deep,faceaudio} render a virtual avatar based on operator-specific geometry. \citet{lombardi2018deep} propose to learn a Variational Autoencoder (VAE) with an encoder that predicts a viewpoint-independent latent variable and a decoder that can be conditioned with extrinsic and intrinsic variables controlling the camera pose, speech, identity, and gaze.
\citet{vr_facial} extend this idea and generate ground truth data with expression-preserving style transfer networks, which map between the VR camera domain and the avatar domain.
More recently, \citet{faceaudio} predict facial coefficients that parametrize a codec avatar with audio and gaze information only. This is especially useful if the lower face region is occluded as well, e.g. with a medical mask.
In contrast to these methods, our pipeline does not assume pre-trained personalized parametric face models and can therefore be adapted to a specific operator with much less preparation time.

\paragraph{Neural Rendering}
Very recently, \citet{face_nerf} proposed to learn facial image animation using dynamic neural radiance fields~\cite{nerf}. Whereas they generate impressive results, their pipeline must be trained for a specific operator, which is impracticable in our application due to the long training time.

\section{Method}

\begin{figure}
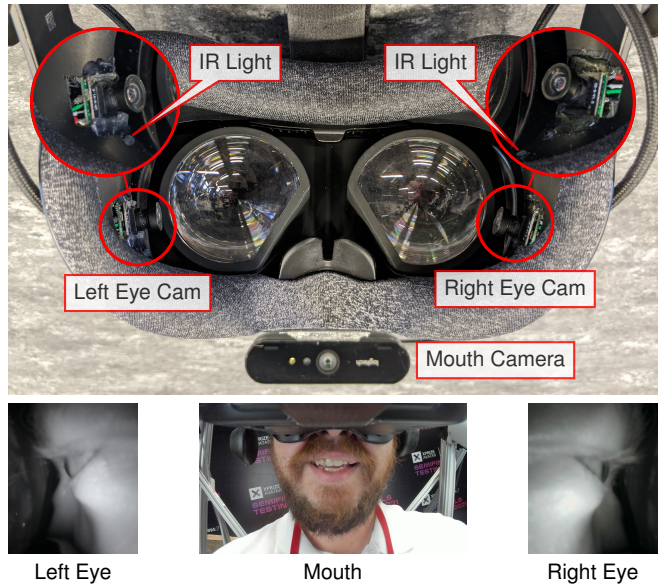


\begin{tikzpicture}[
	font=\footnotesize\sffamily,
	line/.style={red, line width=1pt},
	box/.style={rectangle,draw=red,inner sep=0.3333em,thick,align=center,text height=1.3ex,text depth=.25ex,text centered,fill=white,opacity=0.80}]

\begin{pgfonlayer}{background}
\node[anchor=south west,inner sep=0] (img) at (0,0) {\includegraphics[width=\linewidth,angle=180,clip,trim=0 400 0 200]{images/vr.jpg}};
\end{pgfonlayer}
\node[box] at (rel cs:x=75,y=9,name=img) {Mouth Camera};

\node[circle,minimum width=1cm,draw=red,very thick] (right_eye) at (rel cs:x=78,y=44,name=img) {};
\node[box,below=0.1 of right_eye] {Right Eye Cam};
\begin{scope}[shift={(rel cs:x=85,y=75,name=img)}]
  \clip (0,0) circle (1.0cm);
  \node[rotate=220] at (0.2,0.29) {\includegraphics[width=3.0cm]{images/right_eye_cam.jpg}};
  \node[circle,minimum width=1.96cm,draw=red,very thick] (right_eye_circle) at (0,0) {};
\end{scope}
\draw[red,thick] (right_eye) -- (right_eye_circle);

\node[circle,minimum width=1cm,draw=red,very thick] (left_eye) at (rel cs:x=20,y=43,name=img) {};
\node[box,below=0.1 of left_eye] {Left Eye Cam};
\begin{scope}[shift={(rel cs:x=15,y=75,name=img)}]
  \clip (0,0) circle (1.0cm);
  \node[rotate=160] at (0,0) {\includegraphics[width=3.0cm]{images/left_eye_cam.jpg}};
  \node[circle,minimum width=1.96cm,draw=red,very thick] (left_eye_circle) at (0,0) {};
\end{scope}

\draw[red,thick] (left_eye) -- (left_eye_circle);

\node[box,rectangle callout,callout absolute pointer={(rel cs:x=19.2,y=66.5,name=img)}] at (rel cs:x=35,y=85,name=img) {IR Light};
\node[box,rectangle callout,callout absolute pointer={(rel cs:x=78.5,y=62.7,name=img)}] at (rel cs:x=65,y=85,name=img) {IR Light};

\node[anchor=north west,inner sep=0,label=south:Left Eye] at ($(img.south west)+(0,-0.1)$) {\includegraphics[height=2cm,clip,trim=0 100 0 100]{images/vr_cams/left.png}};
\node[anchor=north east,inner sep=0,label=south:Right Eye] at ($(img.south east)+(0,-0.1)$) {\includegraphics[height=2cm,clip,trim=0 100 0 100]{images/vr_cams/right.png}};
\node[anchor=north,inner sep=0,label=south:Mouth] at ($(img.south)+(0,-0.1)$) {\includegraphics[height=2cm]{images/vr_cams/mouth.png}};
\end{tikzpicture}
\vspace{-.6cm}
\caption{The modified Valve Index VR headset. We attached three additional cameras to capture the eyes and the mouth expression of the operator. The inside of the VR headset is lit using IR LEDs. We show the corresponding camera views at the bottom.}
\label{fig:index} \vspace{-1ex}
\end{figure}

\subsection{Avatar System and Modified VR Headset}

Our robotic avatar system is described in \citep{schwarz2021nimbro}. Briefly put,
it allows a human operator to immersive themselves into a remote robot and to interact and cooperate with humans at the remote site.

For visualization on the operator side, we use the Valve Index VR headset. In order to capture the mouth expression, we attach a Logitech Brio webcam below the HMD (see \cref{fig:index}). Furthermore, we allow for eye tracking by mounting two additional cameras and IR LEDs inside the HMD.

\begin{figure*}
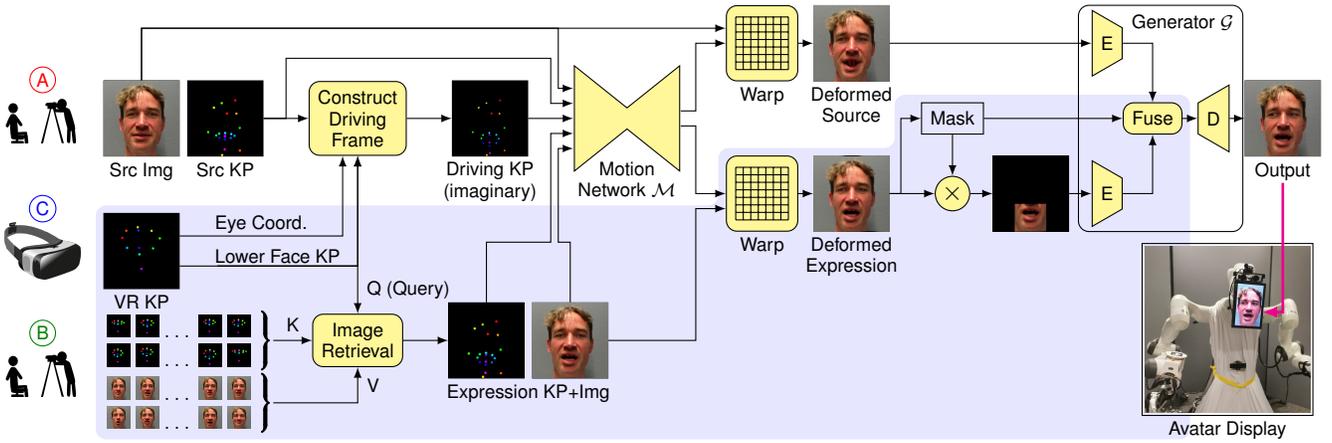

	\centering
	\begin{tikzpicture}[
	font=\sffamily\scriptsize,
	inp/.style={draw,minimum width=1cm,minimum height=1cm,inner sep=0},
	m/.style={draw,rounded corners,fill=yellow!50,align=center},
	node distance=0.6,
	every label/.style={font=\sffamily\scriptsize,align=center,inner sep=2pt,text depth=1pt},
	label distance=0pt,
	]
	
	\node[inp,label=south:Src Img] (source_img) {\includegraphics[width=1cm]{images/inference_figure/sources/mischa4.png}};

	\node[inp,right=0.1 of source_img,label=south:Src KP] (source_kp) {\includegraphics[width=1cm]{images/inference_figure/16460712564617816kp_source_black.png}};
	
	\node[inner sep=0,left=.3cm of source_img,label={[red,draw,circle,text depth=0,inner sep=1pt]north:A}] (portrait) {
		\includegraphics[width=1cm,clip,trim=20 0 20 0]{images/portrait.png}
	};
	
	\node[m,right=of source_kp] (constr) {Construct\\Driving\\Frame};
	
	\node[inp,right=of constr,label={south:Driving KP\\(imaginary)}] (driving_kp) {\includegraphics[width=1cm]{images/inference_figure/16460712564617816kp_driving_black.png}};
	
	\node[inner sep=0,right=of driving_kp,label={[yshift=0.6em]south:Motion\\Network $\mathcal{M}$}] (motion) {\tikz[scale=0.7]{
			\draw[fill=yellow!50] (-1,-1) -- (-1,1) -- (0,0.2) -- (1,1) -- (1,-1) -- (0,-0.2) -- cycle;}
	};

	\node[m,anchor=west,minimum height=1cm,label=south:Warp] at ($(motion.east)+(0.6,-1)$) (warp_expr) {\tikz{
			\draw[step=0.1,black,thin] (0.0,0.0) grid (0.7,0.7);
	}};
	\node[inp,right=0.2 of warp_expr,label=south:Deformed\\Expression] (deformed_expr) {\includegraphics[width=1cm]{images/inference_figure/16460712564617816expression_deformed.png}};
	
	\node[m,circle,font=\normalsize,inner sep=1pt,right=of deformed_expr] (expr_mult) {$\times$};
	\node[draw] at (expr_mult|-motion) (mask) {Mask};
	
	\node[inp,right=0.3 of expr_mult] (masked) {\includegraphics[width=1cm]{images/inference_figure/16460712564617816expression_deformed_crop3.png}};
	
	\node[draw,fill=yellow!50,trapezium,shape border rotate=270,right=0.3 of masked] (expr_enc) {E};

	\node[m,anchor=west,minimum height=1cm,label=south:Warp] at ($(motion.east)+(0.6,1)$) (warp_source) {\tikz{
			\draw[step=0.1,black,thin] (0.0,0.0) grid (0.7,0.7);
	}};
	\node[inp,right=0.2 of warp_source,label=south:Deformed\\Source] (deformed_source) {\includegraphics[width=1cm]{images/inference_figure/16460712564617816source_deformed.png}};
	
	\node[draw,fill=yellow!50,trapezium,shape border rotate=270,anchor=west] at (deformed_source-|expr_enc.west) (source_enc) {E};

	\node[m,anchor=west] at ($(expr_enc.east)!0.5!(source_enc.east)+(0.0,0)$) (fuse) {Fuse};
	\node[draw,fill=yellow!50,trapezium,shape border rotate=90,right=0.2 of fuse] (dec) {D};
	\node[inp,right=0.2 of dec,label=south:Output] (out) {\includegraphics[width=1cm]{images/inference_figure/16460712564617816pred.png}};

	\node[fit=(fuse)(expr_enc)(source_enc)(dec),draw=black,rounded corners,inner sep=5pt] (g) {};
	\node[anchor=north east,align=right] at (g.north east) {Generator $\mathcal{G}$};

	\node[inp,below=0.75cm of source_img,label=south:VR KP] (vr_kp) {\includegraphics[width=1cm]{images/inference_figure/16460712564617816kp_vr_black.png}};
	\coordinate (projin) at ($(vr_kp.east)+(0,-0.2)$);
	\coordinate[m,right=0.5 of projin] (project); %
	
	\node[inner sep=0,left=.3cm of vr_kp,label={[blue,draw,circle,text depth=0,inner sep=1pt]north:C}] {
		\includegraphics[width=1cm]{images/vr_headset.png}
	};

	\node[m,below=2.1cm of constr] (retrieval) {Image\\Retrieval};
	
	\node[anchor=west,matrix,matrix of nodes,inner sep=0,nodes={inner sep=1pt,minimum width=0.4cm,text depth=-5pt},row sep=0.5pt, column sep=-.5pt] at (vr_kp.west|-retrieval.west) (keys) {
		{\includegraphics[width=0.3cm, clip,trim=2.5cm 0 2.5cm 5cm]{images/inference_figure/sources/kp_vr_black_19.png}} & {\includegraphics[width=0.3cm, clip,trim=2.5cm 0 2.5cm 5cm]{images/inference_figure/sources/kp_vr_black_26.png}}  &
		{$\dots$}& {\includegraphics[width=0.3cm, clip,trim=2.5cm 0 2.5cm 5cm]{images/inference_figure/sources/kp_vr_black_61.png}} &
		{\includegraphics[width=0.3cm, clip,trim=2.5cm 0 2.5cm 5cm]{images/inference_figure/sources/kp_vr_black_117.png}} \\
		{\includegraphics[width=0.3cm, clip,trim=2.5cm 0 2.5cm 5cm]{images/inference_figure/sources/kp_vr_black_131.png}} & {\includegraphics[width=0.3cm, clip,trim=2.5cm 0 2.5cm 5cm]{images/inference_figure/sources/kp_vr_black_145.png}}  &
		{$\dots$}& {\includegraphics[width=0.3cm, clip,trim=2.5cm 0 2.5cm 5cm]{images/inference_figure/sources/kp_vr_black_302.png}} & {\includegraphics[width=0.3cm, clip,trim=2.5cm 0 2.5cm 5cm]{images/inference_figure/sources/kp_vr_black_326.png}} \\
	};
	\draw[decorate, decoration = {calligraphic brace},ultra thick] ($(keys.north east)+(0.1,0)$) -- ($(keys.south east)+(0.1,0)$);
	
	\node[anchor=west,matrix,matrix of nodes,inner sep=0,nodes={inner sep=1pt,minimum width=0.4cm,text depth=-5pt},row sep=0.5pt, column sep=-.5pt] at ($(vr_kp.west|-retrieval.west)+(0,-0.83)$) (values) {
		{\includegraphics[width=0.3cm]{images/inference_figure/sources/image_19.png}} & {\includegraphics[width=0.3cm]{images/inference_figure/sources/image_26.png}} &
		{$\dots$}& {\includegraphics[width=0.3cm]{images/inference_figure/sources/image_61.png}} & {\includegraphics[width=0.3cm]{images/inference_figure/sources/image_117.png}} \\
		{\includegraphics[width=0.3cm]{images/inference_figure/sources/image_131.png}} & {\includegraphics[width=0.3cm]{images/inference_figure/sources/image_145.png}}  &
		{$\dots$}& {\includegraphics[width=0.3cm]{images/inference_figure/sources/image_302.png}} & {\includegraphics[width=0.3cm]{images/inference_figure/sources/image_326.png}} \\
	};
	\draw[decorate, decoration = {calligraphic brace},ultra thick] ($(values.north east)+(0.1,0)$) -- ($(values.south east)+(0.1,0)$);
	
	\coordinate (storage) at ($(keys.west)!0.5!(values.west)$);
	\node[inner sep=0,left=.3cm of storage,label={[green!50!black,draw,circle,text depth=0,inner sep=1pt]north:B}] (portrait2) {
		\includegraphics[width=1cm,clip,trim=20 0 20 0]{images/portrait.png}
	};
	
	\node[inp,right=of retrieval] (retrieved_kp) {\includegraphics[width=1cm]{images/inference_figure/16460712564617816kp_expression_black.png}};
	\node[inp,right=0.1 of retrieved_kp] (retrieved_img){\includegraphics[width=1cm]{images/inference_figure/16460712564617816expression.png}};
	\node[every label,anchor=north] at ($(retrieved_kp.south)!0.5!(retrieved_img.south)$) {Expression KP+Img};
	
	\node[draw,inner sep=1pt,anchor=north east,label=south:Avatar Display] (display) at ($(out.south east)+(-0.1,-1.15)$)
	{\includegraphics[width=2.2cm,clip,trim=0 240 50 160]{images/mischa_avatar.jpg}};

	\begin{scope}[-latex]
	\draw (source_img.north) |- ($(warp_source.west)+(0,0.2)$);
	\draw (source_img.north) -- ($(source_img.north|-warp_source.west)+(0,0.2)$) -| ($(motion.west)+(-0.2,0.4)$) -- ($(motion.west)+(0,0.4)$);
	\draw (source_kp) -- (constr);
	\draw (source_kp.east) -- ++(0.3,0) -- ++(0,0.8) -| ($(driving_kp.east)!0.5!(motion.west)+(0,0.2)$) -- ($(motion.west)+(0,0.2)$);
	\draw (constr) -- (driving_kp);
	\draw (driving_kp) -- (motion);
	\draw ($(motion.east)+(0,0.1)$) -- ++(0.2,0) |- ($(warp_source.west)+(-0.1,0)$) -- (warp_source.west);
	\draw (warp_source) -- (deformed_source);
	\draw (deformed_source) -- (source_enc);
	\draw (source_enc) -| (fuse);
	
	\draw ($(motion.east)+(0,-0.1)$) -- ++(0.2,0) |- ($(warp_expr.west)+(-0.1,0)$) -- (warp_expr.west);
	\draw (warp_expr) -- (deformed_expr);
	\draw (deformed_expr) -- (expr_mult);
	\draw (deformed_expr.east) -- ++(0.15,0) |- (mask);
	\draw (mask) -- (expr_mult);
	\draw (mask) |- (fuse);
	\draw (expr_mult) -- (masked);
	\draw (masked) -- (expr_enc);
	\draw (expr_enc) -| (fuse);
	
	\draw (fuse) -- (dec);
	\draw (dec) -- (out);
	\draw[magenta, line width=0.35mm] (out.south) ++(0,-0.35) |- (rel cs:x=70,y=60,name=display);

	\draw (vr_kp.east|-project) -| (constr);
	\draw (project) -| (constr) node [pos=0.215,above,inner sep=1pt]{Lower Face KP};
	\draw ($(vr_kp.east)+(0,0.2)$) -| ($(constr.south)+(-0.2,0)$) node [pos=0.25,above,inner sep=1pt] {Eye Coord.};
	\draw (project) -| (retrieval) node [pos=0.75,right] {Q (Query)};
	\draw (keys.east) ++(0.25,0) -- (retrieval) node [midway,above] {K};
	\draw (values.east) ++(0.25,0) -| (retrieval) node [pos=0.75,right] {V};
	\draw (retrieval) -- (retrieved_kp);
	\draw (retrieved_kp.north) -- ++(0,0.8) -| ($(driving_kp.east)!0.5!(motion.west)+(0,-0.2)$) -- ($(motion.west)+(0,-0.2)$);
	\draw (retrieved_img.north) -- ++(0,0.8) -| ($(motion.west)+(-0.2,-0.4)$) -- ($(motion.west)+(0,-0.4)$);
	\draw (retrieved_img) -| ($(warp_expr.west)+(-0.4,-0.2)$) -- ($(warp_expr.west)+(0,-0.2)$);
	\end{scope}

	\begin{pgfonlayer}{background}
	\draw[fill=blue!10,rounded corners,draw=none]
	($(vr_kp.north west)+(-0.1,0.1)$) -|
	($(warp_expr.north west)+(-0.1,0.1)$) -|
	($(mask.north west)+(-0.35,0.1)$) --
	($(mask.north east)+(0.1,0.1)$) --
	($(fuse.north east)+(0.1,0.1)$) |-
	($(values.south west)+(-0.1,-0.1)$) -- cycle;
	\end{pgfonlayer}
	\end{tikzpicture}
	
	\caption{Inference pipeline for VR Facial Animation. We select a still image from a portrait video of the operator shot before the run as source image \textcolor{red}{(A)}. The remaining frames are used as a key-value storage of expression keypoints and
		corresponding images \textcolor{green!50!black}{(B)}.
		The live keypoints measured inside and outside the VR headset \textcolor{blue}{(C)} are then approximately projected to the
		source image frame, where they are used to retrieve the closest expression keypoints and image from the storage.
		The source keypoints, a constructed set of driving keypoints, and the retrieved expression keypoints then
		enter the motion network $\mathcal{M}$, which estimates warping vectors that deform the source image and the expression
		image to match the driving keypoints.
		The two deformed images then enter an encoder-decoder architecture that fuses them and generates the output image.}
	\label{fig:infer}
	\vspace{-1ex}
\end{figure*}

\subsection{Basic Expression Mapping Pipeline}
\begin{figure*}[h]
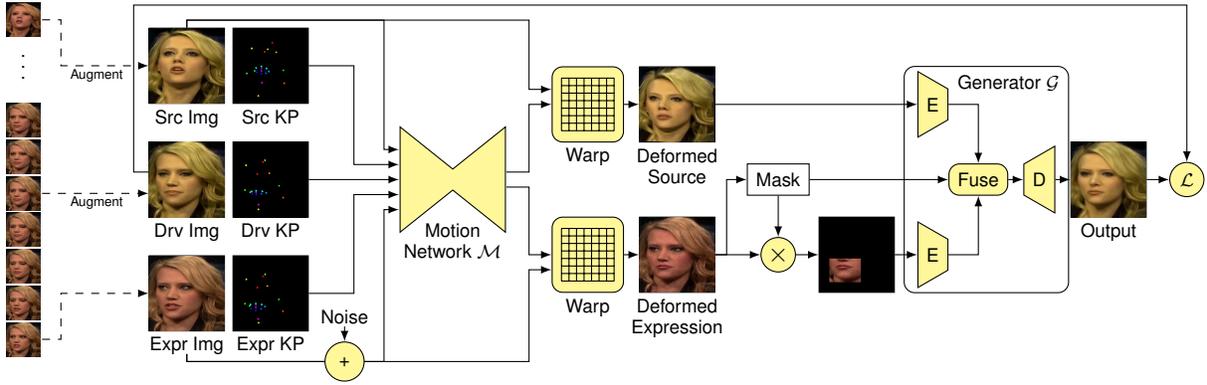

	\centering\newlength{\vidh}\setlength{\vidh}{.45cm}
	\begin{tikzpicture}[
	font=\sffamily\scriptsize,
	inp/.style={draw,minimum width=1cm,minimum height=1cm,inner sep=0},
	m/.style={draw,rounded corners,fill=yellow!50,align=center},
	node distance=0.6,
	every label/.style={font=\sffamily\scriptsize,align=center,inner sep=2pt,text depth=1pt},
	label distance=0pt,
	]

	\node[inp,label=south:Src Img] (source_img) {\includegraphics[width=1cm]{images/train_sequence/a1_source.png}};
	\node[inp,right=0.1 of source_img,label=south:Src KP] (source_kp) {\includegraphics[width=1cm]{images/train_sequence/a2_source_kp_black.png}};

	\node[inp,below=.5 of source_img,label=south:Drv Img] (driv_img) {\includegraphics[width=1cm]{images/train_sequence/a4_driving.png}};
	\node[inp,right=0.1 of driv_img,label=south:Drv KP] (driv_kp) {\includegraphics[width=1cm]{images/train_sequence/a5_driving_kp_black.png}};

	\node[inp,below=.5 of driv_img,label=south:Expr Img] (expr_img) {\includegraphics[width=1cm]{images/train_sequence/a6_expression_no_aug.png}};
	\node[inp,right=0.1 of expr_img,label=south:Expr KP] (expr_kp) {\includegraphics[width=1cm]{images/train_sequence/a7_expression_kp_black.png}};

	\coordinate (inb) at ($(driv_img.west)!0.5!(expr_img.west)$);
	\node[left=1.3 of driv_img,matrix,matrix of nodes,nodes={inner sep=.5pt}] {
	  |(src_src)| \includegraphics[height=\vidh]{images/train_sequence/a1_source_no_aug3.png} \\
	  $\vdots$ \\[.3cm]
	  \includegraphics[height=\vidh]{images/train_sequence/a67_expression_6.png} \\
	  \includegraphics[height=\vidh]{images/train_sequence/a67_expression_5.png} \\
	  |(src_driv)| \includegraphics[height=\vidh]{images/train_sequence/a67_expression_4.png} \\
	  \includegraphics[height=\vidh]{images/train_sequence/a67_expression_3.png} \\
	  \includegraphics[height=\vidh]{images/train_sequence/a67_expression_2.png} \\
	  \includegraphics[height=\vidh]{images/train_sequence/a67_expression_1.png} \\
	  |(src_expr)| \includegraphics[height=\vidh]{images/train_sequence/a67_expression_0.png} \\
	};

	\node[inner sep=0,right=1.2 of driv_kp,label={[yshift=0.6em]south:Motion\\Network $\mathcal{M}$}] (motion) {\tikz[scale=0.7]{
			\draw[fill=yellow!50] (-1,-1) -- (-1,1) -- (0,0.2) -- (1,1) -- (1,-1) -- (0,-0.2) -- cycle;}
	};

	\node[m,anchor=west,minimum height=1cm,label=south:Warp] at ($(motion.east)+(0.6,-1)$) (warp_expr) {\tikz{
			\draw[step=0.1,black,thin] (0.0,0.0) grid (0.7,0.7);
	}};
	\node[inp,right=0.2 of warp_expr,label=south:Deformed\\Expression] (deformed_expr) {{\includegraphics[width=1cm]{images/train_sequence/a8_expression_deformed.png}}};

	\node[m,circle,font=\normalsize,inner sep=1pt,right=of deformed_expr] (expr_mult) {$\times$};
	\node[draw] at (expr_mult|-motion) (mask) {Mask};

	\node[inp,right=0.3 of expr_mult] (masked) {{\includegraphics[width=1cm]{images/train_sequence/a9_expression_deformed_crop.png}}};

	\node[draw,fill=yellow!50,trapezium,shape border rotate=270,right=0.3 of masked] (expr_enc) {E};

	\node[m,anchor=west,minimum height=1cm,label=south:Warp] at ($(motion.east)+(0.6,1)$) (warp_source) {\tikz{
			\draw[step=0.1,black,thin] (0.0,0.0) grid (0.7,0.7);
	}};
	\node[inp,right=0.2 of warp_source,label=south:Deformed\\Source] (deformed_source) {{\includegraphics[width=1cm]{images/train_sequence/a3_source_deformed.png}}};

	\node[draw,fill=yellow!50,trapezium,shape border rotate=270,anchor=west] at (deformed_source-|expr_enc.west) (source_enc) {E};

	\node[m,anchor=west] at ($(expr_enc.east)!0.5!(source_enc.east)+(0.0,0)$) (fuse) {Fuse};
	\node[draw,fill=yellow!50,trapezium,shape border rotate=90,right=0.2 of fuse] (dec) {D};
	\node[inp,right=0.2 of dec,label=south:Output] (out) {{\includegraphics[width=1cm]{images/train_sequence/a10_out.png}}};
	\node[draw,circle,fill=yellow!50,right=0.3 of out,inner sep=2pt] (loss_sub) {$\mathcal{L}$};

	\node[fit=(fuse)(expr_enc)(source_enc)(dec),draw=black,rounded corners,inner sep=5pt] (g) {};
	\node[anchor=north east,align=right] at (g.north east) {Generator $\mathcal{G}$};

	\begin{scope}[-latex]
	\draw[dashed] (src_src) -- ++(0.5,0) |- (source_img) coordinate[pos=1.0] (augmenta);
	\draw[dashed] (src_driv) -- ++(0.5,0) |- (src_driv-|driv_img.west) coordinate[pos=1.0] (augmentb);
	\draw[dashed] (src_expr) -- ++(0.5,0) |- (expr_img);

	\draw (source_img.north) -- ++(0,0.1) -| ($(warp_source.west)+(-0.3,0.2)$) -- ($(warp_source.west)+(0,0.2)$);
	\draw (source_img.north) -- ++(0,0.1) -| ($(motion.west)+(-0.2,0.4)$) -- ($(motion.west)+(0,0.4)$);
	\draw (source_kp.east) -- ++(0.3,0) -| ($(driv_kp.east)!0.5!(motion.west)+(0,0.2)$) -- ($(motion.west)+(0,0.2)$);
	\draw (driv_kp) -- (motion);

	\draw (expr_kp.east) -- ++(0.3,0) -| ($(driv_kp.east)!0.5!(motion.west)+(0,-0.2)$) -- ($(motion.west)+(0,-0.2)$);
	\draw (expr_img.south) ++(0,-0.3) -- ++(0,-0.1) -| ($(warp_expr.west)+(-0.3,-0.2)$) -- ($(warp_expr.west)+(0,-0.2)$);
	\draw (expr_img.south) ++(0,-0.3) -- ++(0,-0.1) -| ($(motion.west)+(-0.2,-0.4)$)
	node[pos=0.4,circle,draw,fill=yellow!50] (noise) {+}
	-- ($(motion.west)+(0,-0.4)$);

	\draw ($(motion.east)+(0,0.1)$) -- ++(0.2,0) -| ($(warp_source.west)+(-0.3,0)$) -- (warp_source.west) ;
	\draw (warp_source) -- (deformed_source);
	\draw (deformed_source) -- (source_enc);
	\draw (source_enc) -| (fuse);

	\draw ($(motion.east)+(0,-0.1)$) -- ++(0.2,0) -| ($(warp_expr.west)+(-0.3,0)$) -- (warp_expr.west);
	\draw (warp_expr) -- (deformed_expr);
	\draw (deformed_expr) -- (expr_mult);
	\draw (deformed_expr.east) -- ++(0.15,0) |- (mask);
	\draw (mask) -- (expr_mult);
	\draw (mask) |- (fuse);
	\draw (expr_mult) -- (masked);
	\draw (masked) -- (expr_enc);
	\draw (expr_enc) -| (fuse);

	\draw (fuse) -- (dec);
	\draw (dec) -- (out);
	\draw (out) -- (loss_sub);
	\end{scope}

	\node[above=0.2 of noise,inner sep=1pt] (noise_in) {Noise};
	\draw[-latex] (noise_in) -- (noise);
	
	\node[font=\sffamily\tiny,anchor=north east,inner sep=1pt] at ($(augmenta)+(-0.3,0)$) {Augment};
	\node[font=\sffamily\tiny,anchor=north east,inner sep=1pt] at ($(augmentb)+(-0.3,0)$) {Augment};

	\draw[-latex] ($(driv_img.west)+(0,0.1)$) -- ++(-0.2,0) |- ($(source_img.north)+(0,0.3)$) -| (loss_sub);

	\end{tikzpicture} \vspace{-1ex}
	\caption{Training the facial animation network from videos.
	The training loss $\mathcal{L}$ is minimized when the network reconstructs the driving image from source image and keypoints, as well as the driving keypoints. The expression image is chosen from a close time interval around the driving image and is available as auxiliary input which is already close to the target expression.} \vspace{-1ex}
	\label{fig:train}
\end{figure*}

We propose a pipeline that allows to map the appearance $\Lambda_S$ from a source image $I_S$ to the expression $\xi_D$ and the head pose $\theta_D$ (motion) present in the driving frame, which may be present as keypoints only (see \cref{fig:train,fig:infer}).

Similar to \citet{fom}, we separate our pipeline into a keypoint detector $\mathcal{K}$, a motion network $\mathcal{M}$, and an image generator $\mathcal{G}$. The motion network $\mathcal{M}$ uses the keypoints extracted by $\mathcal{K}$ to generate a deformation grid $\mathcal{D}_{S\leftarrow D}$ which warps the source image $I_S$ to the head pose and expression of the driving frame $I_D$.

Solely warping is insufficient to generate a realistic output, though. To address this issue, we add a generator network $\mathcal{G}$ which then creates the final output image $I_O$, given the initial motion estimate $\IT{S}{D}$.
For precise architectural information, we refer to \citet{fom}.

Note that there are approaches that forego using a source image by embedding appearance features directly in the network weights~\cite{face_nerf,vr_facial}, however, making only implicit use of the appearance $\Lambda_S$ as encoded in a given source image allows us to generalize to unseen operators. 

\paragraph{Keypoint Detector Network}
Our keypoint detector $\mathcal{K}$ extracts keypoints from the source image $I_S$ and from the driving image $I_D$. We obtain two sequences of $k$ keypoints, respectively:
\begin{alignat}{2}
\mathcal{K}(I_S) &= [\,kp_S^{(1)},kp_S^{(2)},\dots,kp_S^{(k)}\,] \text{ and} \\
\mathcal{K}(I_D) &= [\,kp_D^{(1)},kp_D^{(2)},\dots,kp_D^{(k)}\,].
\end{alignat}

Unlike \citet{fom}, we separate our keypoint detector into two models:
\begin{itemize}
	\item[(i)] One model $\mathcal{K}_{\mathcal{VR}}$ for the expression of the lower face part, and
	\item[(ii)] a global keypoint detector $\mathcal{K}_{\mathcal{F}}$ that has access to images of whole faces, which detects all other keypoints (eyes, head pose, etc.).
\end{itemize}

$\mathcal{K}_{\mathcal{VR}}$ and $\mathcal{K}_{\mathcal{F}}$ are both based on an Hourglass network~\cite{newell2016stacked}.
This separation is necessary because during inference the operator's facial expression, i.e. the driving frame, cannot be reconstructed from a single image due to occlusions by the VR headset.

The global keypoint detector $\mathcal{K}_{\mathcal{F}}$ is trained to detect keypoints, which are primarily encoding the head pose $\theta$ and information about the operator's eyes. We obtain these keypoints from a pretrained first-order-model keypoint detector~\cite{fom}, which was trained in a self-supervised manner.

In contrast, $\mathcal{K}_{\mathcal{VR}}$ is trained in a supervised manner using annotated images from the Vox-Celeb dataset~\cite{vox}. We annotate Vox-Celeb images by cropping the face and extracting keypoints using the method proposed by \citet{face_alignment}. However, we are only interested in the keypoints of the lower face which are visible in our mouth camera (see \cref{fig:index}).
In order to simulate the lower-face image $I_M$ during training, we crop a random quadratic region with the only constraint that all lower-face keypoints must fit into this region. The cropped region is then resized to 128$\times$128 pixels. 
We therefore implicitly train $\mathcal{K}_{\mathcal{VR}}$ to extract keypoints in partially visible faces---as  captured at inference time by the mouth camera.

For simplification, we define 
\begin{align}
\mathcal{K}(I) := \mathcal{K}_{\mathcal{VR}}(I) \oplus \mathcal{K}_{\mathcal{F}}(I)
\end{align}
to be the concatenation $\oplus$ of both keypoint sequences.

\paragraph{Motion Network}
The motion network $\mathcal{M}$ is also based on an Hourglass architecture and produces a deformation of the source frame appearance $\Lambda_S$, represented with the driving frame's facial expression and head pose $(\xi_D,\theta_D)$.
Similar to \citet{fom}, we first create for each keypoint $kp_S^{(i)} \in \mathcal{K}(I_S)$ a shifted source image that aligns $kp_S^{(i)}$ with the corresponding driving keypoint $kp_D^{(i)} \in \mathcal{K}(I_D)$. These $k$ shifted versions are then fed into $\mathcal{M}$ together with the heatmap representation of the driving keypoints.
The motion network then predicts a deformation grid $\mathcal{D}_{S\leftarrow D}$ which can be used to sample the source image deformed to the driving keypoints \IT{S}{D}. For a broader explanation we refer to \citet{fom}.

To enhance the modeling capabilities, \citet{fom} propose to use local affine transformations for each keypoint instead of just shifting. However, this assumes the existence of a complete driving frame. Since our driving frame, i.e. the mouth image, is partially occluded and captured from a different perspective, we rely on a motion network which processes shifted source images and keypoint heatmaps only. This allows us to create imaginary driving frame keypoints with perspective corrections and arbitrary head poses (see \cref{inference_pipeline}).

\paragraph{Generator Network} \label{inference}
The basic generator network has an encoder-decoder architecture and predicts the output image $I_O$, given the deformed source image $\IT{S}{D}$.

\subsection{Inference with VR Headset} \label{inference_pipeline}
In order to animate an operator controlling our avatar, we first need to capture a source image of them without any occlusions. The major challenge during inference is that we do not have access to a driving frame corresponding to a complete face image. Instead, we have to work with a mouth camera and two cameras capturing the eyes. The basic inference pipeline is illustrated in the non-blue area of \Cref{fig:infer}. 

To obtain valid driving keypoints to which we can map the source frame's appearance $\Lambda_S$, we construct the keypoints $\hat{kp}_D^{(i)}\,$$\in$$\,\,\mathcal{K}(\hat{I}_D)$ of an imaginary driving frame $\hat{I}_D$.
The constructed keypoints $\mathcal{K}(\hat{I}_D)$ should encode (i) the same (frontal facing) head pose as in the source image, (ii) the gaze direction and eye openness of the operator, and (iii) the lower-face expression which is captured by the mouth camera.
Note that (ii) and (iii) include nearly all the expressions that we can capture using the mouth and eye cameras.

Taking (i)-(iii) into account, we therefore build the driving keypoints (as illustrated by "Construct Driving Frame" in \Cref{fig:infer}) from
\begin{align}
\mathcal{K}(\hat{I}_D) = \Pi_S (\,\mathcal{K}_{\mathcal{VR}}(I_M)\,) \oplus \rho(\,\mathcal{K}_{\mathcal{F}}(I_S), \hat{kp}_{eye}\,),
\end{align}
where $I_M$ is the mouth image, $\Pi_S(\cdot)$ (see \cref{eq:proj}) maps each lower-face keypoint $kp_M^{(i)} \in \mathcal{K}_{\mathcal{VR}}(I_M)$ into the source image $I_S$, and
$\rho(\cdot)$ replaces the eye key points detected in $I_S$ with the modified values $\hat{kp}_{eye}$ in order to include the operator's current gaze direction and eye openness.
Furthermore, keypoints which encode the face pose are simply copied from the source image. This is sufficient in our use case, since we move the avatar's head display following the operator's head motions (see \Cref{fig:teaser}).

\subsection{Learned Keypoint Mapping $\Pi_S$} \label{sec:mapping}
Constructing the keypoints of our imaginary driving frame $\mathcal{K}(\hat{I}_D)$ requires mapping keypoints from the mouth camera space to the source frame. A na\"ive approach would be to simply perform a translation and scale adjustment~\cite{schwarz2021nimbro}, however, this does not consider perspective differences. One could also attempt to estimate the transformation matrix that maps from the mouth camera to the source camera. This would force us to estimate exact depth in the mouth image, which is not very robust. Furthermore, there are facial distortions caused by the weight of the HMD, which would not be modeled by such an approach.

Instead, we propose to learn a separate homogeneous transformation matrix $\Ti{M}{S}{(i)}$ that maps from each lower-face keypoint $kp_M^{(i)}$ to the corresponding keypoint $\hat{kp}_D^{(i)}$ in the source frame head pose.
Estimating this transformation is not trivial, since the lower face keypoints exhibit a high variance and learning such a mapping requires corresponding mouth and source image pairs. We therefore capture not just a single source image of the operator, but a whole video sequence including mouth movements and a similar second video when the operator wears the VR headset.
Note that finding corresponding pairs is challenging since we cannot capture both videos simultaneously.

Because establishing correspondences manually is expensive, we solve this alignment problem approximately by (i) capturing two videos with roughly the same mouth expressions and (ii) iteratively refining the learned keypoint transformations based on the currently associated pairs.
This iterative process has the following steps:
\begin{enumerate}
	\item Extract the keypoints of all source and mouth video frames.
	\item Initialize each homogeneous transformation $\Ti{M}{S}{(i)}$ from the mean scale difference between the two sets of keypoints.
	\item Map the keypoints of each mouth image into each source frame.
	\item
	For each mapped keypoint sequence, search for the best corresponding source image, yielding $N$ pairs of images.
	\item Optimize each $\Ti{M}{S}{(i)}$ to minimize the Euclidean distance of the current keypoint pairs and goto step~3) unless the maximum number of 1000 iterations is met.
\end{enumerate}
Note that the optimization runs in approx. 10\,s for a reasonable number of 250 calibration images.
In order to create robustness against head movements while capturing the source video and changes of the VR headset relative to the operator,
we define the mapping in a coordinate system relative to the centroid of $I_M$ and $I_S$:

\begin{align} \label{eq:proj}
\Pi_S(\,kp_{M}^{(i)}\,) = \Ti{M}{S}{(i)} \,(\,kp_M^{(i)}-\overline{\mathcal{K}_{\mathcal{VR}}}(I_M)\,) + \overline{\mathcal{K}_{\mathcal{VR}}}(I_S),
\end{align}
where $\overline{\mathcal{K}_{\mathcal{VR}}}(I_S), \overline{\mathcal{K}_{\mathcal{VR}}}(I_M)$ are the mean values of the lower-face keypoints in the source image and the VR camera mouth image, respectively.
As demonstrated in \Cref{fig:qualitative}, this normalization also generates robustness to switching the operator, who is controlling our system, at inference time.

\subsection{Extended Inference Pipeline with Image Retrieval} \label{extended}
Whereas the head and eye positions can be encoded well using only keypoints, it is highly challenging to generate proper mouth expressions using only one source image and few keypoints. To address this issue, we capture not just a single source image but multiple source frames in a video. This allows us to dynamically change the source image to the one which is closest to the projected lower-face keypoints of the current mouth image. 
In such an approach, the pipeline would then need to apply only small corrections to the facial expression and could, thus, generate more accurate mouth expressions. However, in this na\"ive setup, non-negligible flickering effects would appear whenever we change the source view.

Instead, we propose to train a modified generator network that allows to decode the information of not just one source frame but also the lower-face information of a second source frame, which should be closer to the target. Therefore, the primary source frame $I_S$ always remains constant while the second source frame, which we call the expression frame $I_E$, can change arbitrarily (see \Cref{fig:infer}). This has a positive effect on temporal continuity and counteracts flickering during source frame changes significantly.

Following standard information retrieval, we compare the current query $$Q=\mathcal{K}_{\mathcal{VR}}(I_M)$$ with all keys
$$K = [\,\mathcal{K}_{\mathcal{VR}}(I_{S_1}),\dots, \mathcal{K}_{\mathcal{VR}}(I_{S_n})\,]$$
via
$$ \sum_l ||\Pi_{S_i}(Q^{(l)})-k_{i}^{(l)}||_2,\, k_i \in K$$
to retrieve the optimal index of the (image, keypoints) tuples
$$V = [\,(\,I_{S_1},\mathcal{K}({I_{S_1}})\,),\dots, (\,I_{S_n},\mathcal{K}({I_{S_n}})\,)\,].$$

We modify the pipeline and generator network accordingly:
\begin{itemize}
	\item[(i)] Use the motion network to generate a deformed image of both the source frame and the expression frame ($\IT{S}{D},\IT{E}{D}$).
	\item[(ii)] Split the generator into two separate encoder networks, where the first encoder $\mathcal{G}^{Enc}_S$ extracts the source image features
	\begin{align} \f{S}{D} = \mathcal{G}^{Enc}_S(\IT{S}{D})
	\end{align}
	and the second encoder $\mathcal{G}^{Enc}_E$ extracts the expression image features
	\begin{align}
	\f{E}{D} = \mathcal{G}^{Enc}_E(m \odot \IT{E}{D}),
	\end{align}
	where $m \in $\{$0,1$\}$^{N\times N}$ is a mask that hides all information except the lower face region of $\IT{E}{D}$.
	\item[(iii)] Fuse the activation by
	\begin{align}
	\medmath {\f{S,E}{D} = \frac{m_\downarrow}{2}\odot(\f{S}{D} + \f{E}{D}) + (1-m_\downarrow)\odot \f{S}{D}},
	\end{align}
	where $\odot$ denotes element-wise multiplication and $m_\downarrow$ is a down-scaled version of the binary mask $m$.
	\item[(iv)] Decode the fused representation with the decoder $\mathcal{G}^{Dec}$ of the generator network 
	\begin{align}
	I_O = \mathcal{G}^{Dec}(\f{S,E}{D}).
	\end{align}
\end{itemize}
This additional branch to our inference pipeline is visualized in the blue-marked area of \Cref{fig:infer}.

When capturing the two videos (with and without HMD), it is important to have a high variety of mouth expression in the set of source images.
We therefore propose to capture two videos of the operator, while reading a sentence that covers many phonemes. The sentence "That quick beige fox jumped in the air over each thin dog, look out he shouts for he's foiled you again, creating chaos", is known to match these requirements and gave us good results during testing.

\subsection{Training}

\noindent
The training pipeline is illustrated in \Cref{fig:train}.

During inference, we use information retrieval to select the current expression frame, based on the optimized keypoint transformations \T{M}{S}. 
During training, however, we use the Vox-Celeb dataset~\cite{vox}, which mainly consists of celebrities being interviewed. In this setup, it is not trivial to select an expression frame: If we would just set the expression image to be the driving image itself, the network would learn to simply copy the information.

To avoid this, we select the expression frame from a small interval around the driving image. This makes the assumption that temporally close frames also exhibit similar expressions.

The generator network (especially the feature encoder sub-modules) can thus learn to mostly ignore the lower-face region of the motion-transmitted source image, since the expression image is generally closer to the driving frame. This, however, leads to temporal instabilities whenever the expression frame is switched. We counteract this by choosing a reasonable interval around the driving frame and augment the chosen expression frame using color jittering and injection of several types of random noise as proposed by \citet{noise}.
We argue that in this setup, the generator network is now explicitly guided to keep the mapped source frame information $\IT{S}{D}$ to generate a proper facial animation.

\subsection{Eye Tracking and Animation}\label{sec:eye_tracking}

\begin{figure}
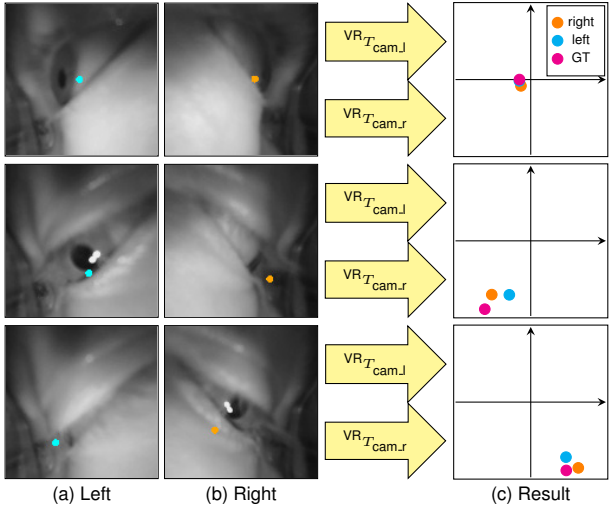

 \centering\newlength{\eyeh}\setlength{\eyeh}{2cm}
 \begin{tikzpicture}[font=\sffamily\footnotesize,arr/.style={single arrow,fill=yellow!50,font=\tiny,minimum height=1.6cm,inner sep=3pt}]
  \node[matrix, matrix of nodes, nodes={inner sep=0.5pt,draw,anchor=center}, row sep=2pt, column sep=2pt] {
    \includegraphics[angle=90,origin=c,height=\eyeh] {images/eye_tracking/left0009.png} &
    \includegraphics[angle=-90,origin=c,height=\eyeh] {images/eye_tracking/right0009.png} &
    |[draw=none]| \tikz{
       \node[arr] (a) {$\T{\text{cam\_l}}{\text{VR}}$};
       \node[arr,below=.5 of a] (b) {$\T{\text{cam\_r}}{\text{VR}}$};
    } &
    |(res)| \tikz{
      \begin{axis}[ymin=-1.1,ymax=1.1,xmin=-1.1,xmax=1.1, scale only axis, width=\eyeh, height=\eyeh, axis equal,xtick=\empty,ytick=\empty,
      axis lines=center,%
      every axis legend/.append style={inner sep=2pt,font=\tiny,row sep=-2pt},legend to name=eye_legend]
       \addplot[orange,only marks] table [x index=4, y index=5] {images/eye_tracking/data0009.txt};
       \addlegendentry{right}
       \addplot[cyan,only marks]    table [x index=0, y expr=\thisrowno{1}-0.01] {images/eye_tracking/data0009.txt};
       \addlegendentry{left}
       \addplot[magenta,only marks] table [x index=2, y index=3] {images/eye_tracking/data0009.txt};
       \addlegendentry{GT}
      \end{axis}
    } \\
    \includegraphics[angle=90,origin=c,height=\eyeh] {images/eye_tracking/left0019.png} &
    \includegraphics[angle=-90,origin=c,height=\eyeh] {images/eye_tracking/right0019.png} &
    |[draw=none]| \tikz{
       \node[arr] (a) {$\T{\text{cam\_l}}{\text{VR}}$};
       \node[arr,below=.5 of a] (b) {$\T{\text{cam\_r}}{\text{VR}}$};
    } &
    \tikz{
      \begin{axis}[ymin=-1.1,ymax=1.1,xmin=-1.1,xmax=1.1, scale only axis, width=\eyeh, height=\eyeh, axis equal,xtick=\empty,ytick=\empty,
      axis lines=center]
       \addplot[orange,only marks] table [x index=4, y index=5]  {images/eye_tracking/data0019.txt};
       \addplot[cyan,only marks]    table [x index=0, y index=1] {images/eye_tracking/data0019.txt};
       \addplot[magenta,only marks] table [x index=2, y index=3] {images/eye_tracking/data0019.txt};
      \end{axis}
    } \\
    \includegraphics[angle=90,origin=c,height=\eyeh] {images/eye_tracking/left0020.png} &
    \includegraphics[angle=-90,origin=c,height=\eyeh] {images/eye_tracking/right0020.png} &
    |[draw=none]| \tikz{
       \node[arr] (a) {$\T{\text{cam\_l}}{\text{VR}}$};
       \node[arr,below=.5 of a] (b) {$\T{\text{cam\_r}}{\text{VR}}$};
    } &
    \tikz{
      \begin{axis}[ymin=-1.1,ymax=1.1,xmin=-1.1,xmax=1.1, scale only axis, width=\eyeh, height=\eyeh, axis equal,xtick=\empty,ytick=\empty,
      axis lines=center]
       \addplot[cyan,only marks]    table [x index=0, y index=1] {images/eye_tracking/data0020.txt};
       \addplot[orange,only marks] table [x index=4, y index=5] {images/eye_tracking/data0020.txt};
       \addplot[magenta,only marks] table [x index=2, y index=3] {images/eye_tracking/data0020.txt};
      \end{axis}
    } \\
    |[draw=none,font=\sffamily\scriptsize]| (a) Left &
    |[draw=none,font=\sffamily\scriptsize]| (b) Right & &
    |[draw=none,font=\sffamily\scriptsize]| (c) Result \\
  };
  \node[anchor=north east,inner sep=1pt] at (res.north east) {\ref{eye_legend}};
 \end{tikzpicture}
 \vspace{-1em}
 \caption{Eye tracking. We show the eye keypoints in the left and right image frame, which are learned without
 direct supervision (a, b). The resulting transformed gaze direction prediction is shown
 in (c) in cyan and orange, with the ground truth in magenta. The transformations $\T{\text{cam}}{\text{VR}}$ are
 learned.}
 \label{fig:eye_tracking}
\end{figure}

We introduce an image-driven eye tracking pipeline that needs to be calibrated once and can be trained in less than a minute for an operator. To obtain training data, we request the operator to follow with their gaze a red dot that moves in the VR display. This gives us calibration/training triplets of left eye images, right eye images and 3D gaze directions.

\paragraph{Network Architecture}
We build a very lightweight hourglass network with only two downsampling and two upsampling layers that takes an input eye image and outputs a heatmap which is used to generate a single keypoint (see \cref{fig:eye_tracking}). We map this keypoint coordinate into the VR space using a learned homogeneous transformation $\T{\text{cam}}{\text{VR}}$, which is jointly optimized with the hourglass network. While the homogeneous transformation is trained with supervision, we train the hourglass end-to-end in a self-supervised manner.

\paragraph{Inference} \label{sec:eye_infer}
During inference, we take the mean prediction $p$ of both the left eye prediction $p_L \in [-1,1]^2$ and the right eye prediction $p_R \in [-1,1]^2$. We, furthermore, estimate a normalized confidence measure
\begin{align}
c = 1- \frac{1}{2 \sqrt{2}} ||p_L-p_R||_2 \in [0,1]
\end{align}
which is large when both predicted eye coordinates are close to each other.

\begin{figure}
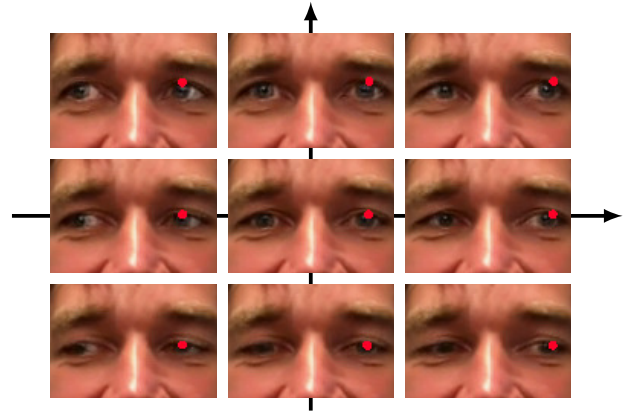

	\centering\newlength{\eyew}\setlength{\eyew}{2.2cm}
	\begin{tikzpicture}
	 \draw[-latex, line width=0.5mm] (-.46\linewidth,0) -- (.48\linewidth,0);
	 \draw[-latex, line width=0.5mm] (0,-.3\linewidth) -- (0,.33\linewidth);
	
	 \node[matrix,matrix of nodes,nodes={inner sep=2.2pt}] {
	   \includegraphics[width=\eyew,clip,trim=2.6cm 3.8cm 2.6cm 2.6cm]{images/eye_results_interactive_tool/16459078989953194.png}
		&
		\includegraphics[width=\eyew,clip,trim=2.6cm 3.8cm 2.6cm 2.6cm]{images/eye_results_interactive_tool/16459079836553846.png}
		&
		\includegraphics[width=\eyew,clip,trim=2.6cm 3.8cm 2.6cm 2.6cm]{images/eye_results_interactive_tool/16459079716350594.png} \\
		\includegraphics[width=\eyew,clip,trim=2.6cm 3.8cm 2.6cm 2.6cm]{images/eye_results_interactive_tool/16459079097360472.png}
		&
		\includegraphics[width=\eyew,clip,trim=2.6cm 3.8cm 2.6cm 2.6cm]{images/eye_results_interactive_tool/16459078330185142.png}
		&
		\includegraphics[width=\eyew,clip,trim=2.6cm 3.8cm 2.6cm 2.6cm]{images/eye_results_interactive_tool/16459078422300530.png} \\
		\includegraphics[width=\eyew,clip,trim=2.6cm 3.8cm 2.6cm 2.6cm]{images/eye_results_interactive_tool/16459079165870170.png}
		&
		\includegraphics[width=\eyew,clip,trim=2.6cm 3.8cm 2.6cm 2.6cm]{images/eye_results_interactive_tool/16459079211364978.png}
		&
		\includegraphics[width=\eyew,clip,trim=2.6cm 3.8cm 2.6cm 2.6cm]{images/eye_results_interactive_tool/16459079310476552.png}
		\\
	 };
	\end{tikzpicture} \vspace{-2ex}
	\caption{Illustration of the eyes reacting to keypoint manipulations. The center image defines the origin and the other images define the boundaries of the normalized eye coordinate system.}
	\label{fig:gaze}
\end{figure}

At this stage, only the recognition of the eye openness remains. We found that the eye openness strongly correlates with the gaze direction, i.e. the more an operator looks down the less open the eyes are.
This property was also learned by our networks. To simulate both eyes, the networks only use one keypoint at the upper eyelid directly above the pupil center of the right eye (see \cref{fig:gaze}).
During inference, we can thus control the gaze and eye openness by modifying a single keypoint of the source image.

The assumptions above apply as long as the eyes are not completely closed, i.e. not when blinking.
It is beneficial to detect this case without requiring additional annotations.
Our experiments showed that whenever the eyes are closed, there is a very low confidence value.
This is due to the fact that the predictions of both eyes are very different, since we did not equip the networks with the capabilities of handling such cases.
Hence, we detect closed eyes whenever the confidence parameter is below a threshold $\lambda_{C}$. This also
has the benefit of hiding implausible eye configurations from the viewer by showing the operator with closed eyes.

\paragraph{Eye Coordinate System}
One remaining problem is still to define the region in the source image in which the eye keypoint can move, i.e. a coordinate system mapping.
One possibility is to automatically find these boundaries. However, this requires to capture a second video of the operator (without VR headsets) in which the eyes are moving.

To decrease required capture time, we built an interactive annotation tool, which renders the source view with eye keypoints at the current cursor position. This allows us to manually define the boundaries of the possible eye keypoints in the source image, which are illustrated in \Cref{fig:gaze}. The annotated boundaries then define a normalized coordinate system which is centered in the frontal-facing gaze direction.

\subsection{Temporal Consistency Filters}

We apply several filters to the image retrieval process and facial keypoints to enhance the temporal consistency.

\paragraph{Expression Frame Filtering}
We apply a straightforward filter that regulates the expression frame retrieval process:
We only allow to change the expression frame $I_E$, when there is another frame 1+$\lambda_{swap}$ times closer to the current projected lower-face keypoints. This hysteresis avoids fast switching.

We also apply a recursive low-pass filter parameterized by $\lambda_E, \lambda_{\tilde{E}}, \lambda_O$. Instead of simply using the raw expression frame $I_E$, %
we propose to build the current expression frame $I^{(t)}_{\tilde{E}}$ at time step $t$ according to
\begin{align}
I^{(t)}_{\tilde{E}} = \lambda_E I_E^{(t)} + \lambda_{\tilde{E}} \Ii{\tilde{E}}{E}{(t-1)} +\lambda_O \Ii{O}{E}{(t-1)},
\end{align}
where $I_E^{(t)}$ is the current (raw) expression frame, $\Ii{\tilde{E}}{E}{(t-1)}$ is the last expression frame $I^{(t-1)}_{\tilde{E}}$ (recursively) mapped to the current expression frame using the motion network, and $\Ii{O}{E}{(t-1)}$ is the last prediction $I^{(t-1)}_O$ mapped to the current expression frame, respectively. In practice, we choose \\$\lambda_E=0.7$, $\lambda_{\tilde{E}}=0.1$, and $\lambda_O=0.2$. %

\paragraph{Eye Keypoint Filtering}
As explained in \cref{sec:eye_tracking}, we can obtain a confidence value of the current eye position by comparing the left and right eye.
We then recursively low-pass filter the eye position and parametrize the filter with the confidence $c \in [0,1]$. We propose to derive the filtered eye position $\tilde{p}$ from
\begin{align}
\tilde{p}^{(t)} = \lambda_{G}\,p^{(t)} + (1-\lambda_G)\,\tilde{p}^{(t-1)},
\end{align}
where we determine the contribution $\lambda_G \in [0,1]$ of the current eye position estimate $p^{(t)}$ according to:
\begin{align}
\lambda_{G}=
\begin{cases}
5c -4,& \text{ if } c \geq 0.8\\
0, &\text{ otherwise.}
\end{cases}
\end{align}
This low-pass filter is especially useful when the prediction~$p$ of the eye tracking network is noisy.
\begin{figure}
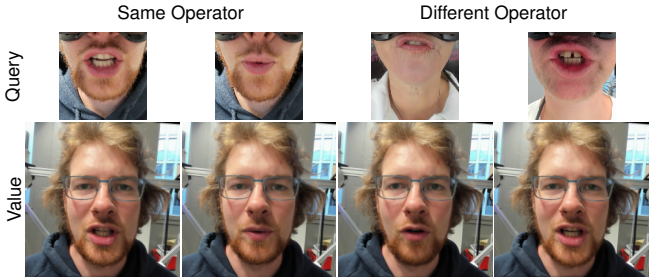

  \centering
  \begin{tikzpicture}[font=\sffamily\scriptsize]
    \node[inner sep=0,matrix, matrix of nodes, nodes={inner sep=0,anchor=center}, row sep=1pt, column sep=1pt] (m) {
      |[rotate=90]| Query &
      \includegraphics[width=0.065\textwidth]{images/qualitative_results_chris/retrieval/1/16458898096623292_vr.png} &
      \includegraphics[width=0.065\textwidth]{images/qualitative_results_chris/retrieval/2/16458898070162890_vr.png} &
      \includegraphics[width=0.065\textwidth]{images/qualitative_results_chris/retrieval/4/16458013006251466_vr.png} &
      \includegraphics[width=0.065\textwidth]{images/qualitative_results_chris/retrieval/5/16458910246696480_vr.png} \\
      |[rotate=90]| Value &
      \includegraphics[width=0.115\textwidth]{images/qualitative_results_chris/retrieval/1/16458898096623292expression.png} &
      \includegraphics[width=0.115\textwidth]{images/qualitative_results_chris/retrieval/2/16458898070162890expression.png} &
      \includegraphics[width=0.115\textwidth]{images/qualitative_results_chris/retrieval/4/16458013006251466expression.png} &
      \includegraphics[width=0.115\textwidth]{images/qualitative_results_chris/retrieval/5/16458910246696480expression.png} \\
    };
    \coordinate (a) at ($(m-1-2.north)!0.5!(m-1-3.north)$);
    \node[above=.0 of a] {Same Operator};

    \coordinate (b) at ($(m-1-4.north)!0.5!(m-1-5.north)$);
    \node[above=.0 of b] {Different Operator};
  \end{tikzpicture} \vspace{-3.5ex}
	\caption{Image retrieval process experiments. Unlike the first two columns, the last two columns are examples for image retrievals where the operator keys and values are different from the operator in the query.}
	\label{fig:retrieval}
\end{figure}

\newcommand{\eyeplot}[1]{%
  \begin{axis}[
        ymin=-1,ymax=1,xmin=-0.9,xmax=0.9, scale only axis, width=\eyeh, height=\eyeh,
        axis equal,xtick=\empty,ytick=\empty,
        axis lines=center
   ]
   \addplot[red,only marks] table [x index=0, y expr=-\thisrowno{1},row sep=\\] {
        #1 \\
   };
  \end{axis}
}

\newcommand{\kpinput}[2]{%
  \tikz{
    \node[inner sep=0] (a) {\includegraphics[width=0.066\textwidth]{#1}};
    \node[inner sep=0, right=0.05 of a] {\tikz{\eyeplot{#2}}};
  }
}

\begin{figure*}[h]
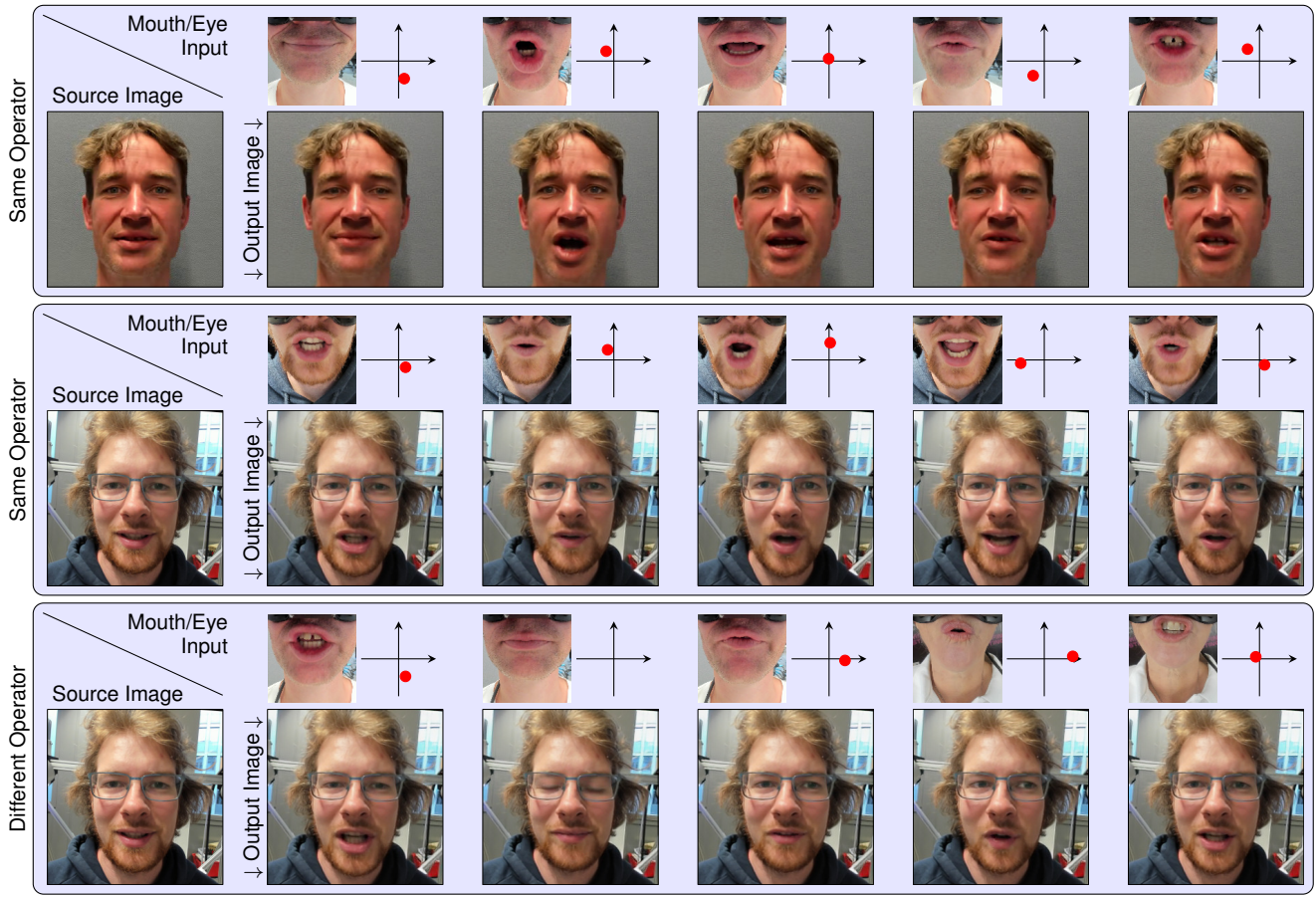
 \centering
	\setlength{\eyeh}{1cm}
	\tikz[font=\sffamily\footnotesize]{
      \node[matrix,matrix of nodes, nodes={inner sep=0,anchor=center},column sep=15pt,row sep=1.5pt] (m) {
		\tikz{
		  \node[align=right] (a) at (0.8,0) {Mouth/Eye\\Input};
		  \node (b) at (0,-0.8) {Source Image};
		  \draw (-1,0.3) -- (1.4,-0.8);
		} &
		\kpinput{images/qualitative_results_mischa/1/16461332258117270_vr.png}{0.14957969 0.46967393} &
		\kpinput{images/qualitative_results_mischa/6/16461456916576848_vr.png}{-0.20597005 -0.2625275} &
		\kpinput{images/qualitative_results_mischa/7/16461456948038404_vr.png}{-0.011877477 -0.06438935} &
		\kpinput{images/qualitative_results_mischa/3/16461450274277802_vr.png}{-0.29697284 0.39011693} &
		\kpinput{images/qualitative_results_mischa/8/16461455054082598_vr.png}{-0.31902504 -0.31861007} \\
		\includegraphics[frame,width=0.132\textwidth]{images/best_pairs/m2m/mischa4.png} &
		|(out1)|\includegraphics[frame,width=0.132\textwidth]{images/qualitative_results_mischa/1/16461332258117270pred.png} &
		\includegraphics[frame,width=0.132\textwidth]{images/qualitative_results_mischa/6/16461456916576848pred.png} &
		\includegraphics[frame,width=0.132\textwidth]{images/qualitative_results_mischa/7/16461456948038404pred.png} &
		\includegraphics[frame,width=0.132\textwidth]{images/qualitative_results_mischa/3/16461450274277802pred.png} &
		\includegraphics[frame,width=0.132\textwidth]{images/qualitative_results_mischa/8/16461455054082598pred.png} \\[.3cm]
		\tikz{
		  \node[align=right] (a) at (0.8,0) {Mouth/Eye\\Input};
		  \node (b) at (0,-0.8) {Source Image};
		  \draw (-1,0.3) -- (1.4,-0.8);
		}&
		\kpinput{images/chris_data/1/16461494756798288_vr.png}{0.18509065 0.1954361} &
		\kpinput{images/chris_data/2/16461494832012944_vr.png}{-0.16620032 -0.27065837} &
		\kpinput{images/chris_data/3/16461494845051240_vr.png}{0.031160913 -0.45839363} &
		\kpinput{images/chris_data/13/16461575778628368_vr.png}{-0.6286006 0.08785686} &
		\kpinput{images/chris_data/5/16461523860543094_vr.png}{0.13482738 0.13287412} \\
		\includegraphics[frame,width=0.132\textwidth]{images/qualitative_results_chris/chris_neu.png} &
		|(out2)|\includegraphics[frame,width=0.132\textwidth]{images/chris_data/1/16461494756798288pred.png} &
		\includegraphics[frame,width=0.132\textwidth]{images/chris_data/2/16461494832012944pred.png} &
		\includegraphics[frame,width=0.132\textwidth]{images/chris_data/3/16461494845051240pred.png} &
		\includegraphics[frame,width=0.132\textwidth]{images/chris_data/13/16461575778628368pred.png} &
		\includegraphics[frame,width=0.132\textwidth]{images/chris_data/5/16461523860543094pred.png} \\[.3cm]
		\tikz{
		  \node[align=right] (a) at (0.8,0) {Mouth/Eye\\Input};
		  \node (b) at (0,-0.8) {Source Image};
		  \draw (-1,0.3) -- (1.4,-0.8);
		} &
		\kpinput{images/others_to_chris_data/5/16461579421946052_vr.png}{0.18440858 0.4696105} &
		\kpinput{images/others_to_chris_data/0/16461579330929482_vr.png}{} &
		\kpinput{images/others_to_chris_data/4/16461579484039724_vr.png}{0.425665 0.043322936} &
		\kpinput{images/others_to_chris_data/13/16461587500463400_vr.png}{0.76098895 -0.066927895} &
		\kpinput{images/others_to_chris_data/7/16461587337920730_vr.png}{-0.0996647 -0.053963616} \\
		\includegraphics[frame,width=0.132\textwidth]{images/qualitative_results_chris/chris_neu.png} &
		|(out3)|\includegraphics[frame,width=0.132\textwidth]{images/others_to_chris_data/5/16461579421946052pred.png} &
		\includegraphics[frame,width=0.132\textwidth]{images/others_to_chris_data/0/16461579330929482pred.png} &
		\includegraphics[frame,width=0.132\textwidth]{images/others_to_chris_data/4/16461579484039724pred.png} &
		\includegraphics[frame,width=0.132\textwidth]{images/others_to_chris_data/13/16461587500463400pred.png} &
		\includegraphics[frame,width=0.132\textwidth]{images/others_to_chris_data/7/16461587337920730pred.png} \\
      };

      \begin{pgfonlayer}{background}
      \node[fill=blue!10,draw,rounded corners,fit=(m-1-1)(m-2-6)] (b1) {};
      \node[fill=blue!10,draw,rounded corners,fit=(m-3-1)(m-4-6)] (b2) {};
      \node[fill=blue!10,draw,rounded corners,fit=(m-5-1)(m-6-6)] (b3) {};
      \end{pgfonlayer}
      \node[rotate=90,anchor=south,inner sep=1pt] at (b1.west) {Same Operator};
      \node[rotate=90,anchor=south,inner sep=1pt] at (b2.west) {Same Operator};
      \node[rotate=90,anchor=south,inner sep=1pt] at (b3.west) {Different Operator};
      
      \node[rotate=90,anchor=south,inner sep=1pt] at (out1.west) {$\downarrow$ Output Image $\downarrow$};
      \node[rotate=90,anchor=south,inner sep=1pt] at (out2.west) {$\downarrow$ Output Image $\downarrow$};
      \node[rotate=90,anchor=south,inner sep=1pt] at (out3.west) {$\downarrow$ Output Image $\downarrow$};
    } \vspace{-2ex}
	\caption{Generated faces. In each box we use the same source image (left) to generate a facial reconstruction given the mouth camera image and eye coordinates. Note how the system matches mouth and eye configurations closely (all rows) and even performs inter-person animation (bottom row).}\vspace{-2ex}
	\label{fig:qualitative} \vspace{-1.5ex}
\end{figure*}

\section{Experiments}
We report several qualitative results and discuss our performance at the ANA Avatar XPRIZE semifinals.

\subsection{Qualitative Results}

\paragraph{VR Facial Animation}
\Cref{fig:qualitative} illustrates exemplary results of our full pipeline when (i) mapping from one operator to the appearance of the same operator and (ii) mapping to the appearance of another operator. Note that our complete forward pass runs with 33\,fps on a single NVIDIA RTX 3090 GPU and an image resolution of 256$\times$256.

\paragraph{Image Retrieval}
\Cref{fig:retrieval} visualizes the accuracy of our proposed keypoint-based image retrieval. Experiments have shown that image retrieval from a set of source images allows us to generate much more accurate mouth expressions compared to just a single source. As shown in \Cref{fig:retrieval}, the image retrieval also works well between different operators. 

\subsection{The ANA Avatar XPRIZE Semifinals}
At the semifinals of the ANA Avatar XPRIZE Competition, three scenarios had to be accomplished with our avatar system by a previously unknown operator. The scenarios were repeated in a second run, with the better score persisting.
Before each run, a preparation time of one hour was allotted in which we could introduce the operator to the system.  We also used this time to prepare the facial animation pipeline, including recording two source videos (with and without the HMD), optimizing the keypoint transformations, calibrating and training the eye tracking pipeline, and annotating eye coordinates using our interactive tool. On average, operator adaptation took us about 15\,minutes, which could mostly be done in parallel to the general operator introduction.
The scenarios all required the operator to interact though the avatar system with another person from the jury, i.e. the recipient.  
For the recipient, facial animation is particularly important, as it helps to convey the focus of attention and the emotions of the operator and confirms the recipients that they are interacting with a real person.

XPRIZE defined rigorous scoring criteria. Specific criteria regarding facial animation were:
\begin{enumerate}
	\item The Recipient was able to identify the remote Operator and felt the Operator was present in space.
	\item The Recipient was able to understand the Operator's emotions through the Avatar.
	\item The Recipient felt a sense of shared experience with the remote Operator.
	\item The Recipient was able to understand the Operator's gestures through the Avatar.
	\item The Operator was able to express their emotions.
\end{enumerate}

Our team NimbRo performed very well at the these and all other criteria and was ranked first in the semifinals with an overall score of 99 out of 100 points. \Cref{fig:teaser} shows our avatar system and facial animations during the challenge.

\section{Conclusions}

We have demonstrated an efficient and real-time capable VR Facial Animation pipeline that generalizes well to operators unknown \textit{a proiri} with a fast adaptation process. We show how to accurately solve the alignment problem between images captured with and without an HMD using learned keypoint transformations and iterative refinement.
Furthermore, we demonstrate how to ensure temporal continuity.
Using the proposed method, our team reached an almost perfect score of 99/100 points and was ranked first in the ANA Avatar XPRIZE semifinals event in Miami with a total of 28 teams.
A useful extension of our method could be to forgo selecting an expression frame, but directly use the mouth image as captured by the mouth camera.

\printbibliography

\end{document}